\documentclass[]{damo}

\usepackage{wrapfig}
\usepackage{tabularx}
\usepackage{textcomp}
\usepackage{stfloats}
\usepackage{url}
\usepackage{verbatim}
\usepackage{graphicx}
\usepackage{titlesec}
\usepackage{tocloft}
\usepackage{adjustbox}
\usepackage{multirow}
\usepackage{pifont}
\usepackage{tikz}
\usepackage{comment}
\usepackage{amsmath,amssymb} 
\usepackage{colortbl}  
\usepackage{color}
\usepackage{booktabs} 
\usepackage{hyperref}
\usepackage{graphicx}    
\usepackage{subcaption} 
\usepackage{booktabs}
\usepackage{amsmath} 
\usepackage{multirow} 
\usepackage{booktabs} 
\usepackage{subcaption} 
\RequirePackage{xspace}
\makeatletter
\DeclareRobustCommand\onedot{\futurelet\@let@token\@onedot}
\def\@onedot{\ifx\@let@token.\else.\null\fi\xspace}

\makeatother

\usepackage{makecell}

\definecolor{adptorange}{RGB}{248, 205, 172}
\definecolor{cmpblue}{RGB}{189, 215, 238}
\definecolor{cmpblue}{RGB}{189, 215, 238}

\definecolor{our_red}{RGB}{232,157,160}
\definecolor{our_blue}{RGB}{136,206,230}
\definecolor{our_orange}{RGB}{246,200,168}
\definecolor{our_green}{RGB}{178,211,164}

\definecolor{attn_code0}{RGB}{247,215,200}
\definecolor{attn_code1}{RGB}{238,169,139}
\definecolor{mlp_code0}{RGB}{204,201,221}
\definecolor{mlp_code1}{RGB}{102,95,153}

\definecolor{token_blue}{RGB}{84, 120, 140}

\definecolor{myMagenta}{rgb}{0.9,0,0.4}

\usepackage{pifont}       
\usepackage{bbding}       
\usepackage{fontawesome}
\usepackage{xspace}

\usepackage{float}

\newlength\savewidth

\newcolumntype{x}[1]{>{\centering\arraybackslash}p{#1pt}}
\newcolumntype{y}[1]{>{\raggedright\arraybackslash}p{#1pt}}
\newcolumntype{z}[1]{>{\raggedleft\arraybackslash}p{#1pt}}

\renewcommand{\paragraph}[1]{\vspace{1mm}\noindent\textbf{#1}}
\usepackage{colortbl}
\usepackage{xcolor}
\usepackage{wrapfig}

\renewcommand{\paragraph}[1]{\vspace{1.25mm}\noindent\textbf{#1}}

\usepackage{algorithm}
\usepackage{listings}

\definecolor{codeblue}{rgb}{0.25, 0.5, 0.5}
\definecolor{codekw}{rgb}{0.35, 0.35, 0.75}
\lstdefinestyle{Pytorch}{
    language = Python,
    backgroundcolor = \color{white},
    basicstyle = \fontsize{9pt}{8pt}\selectfont\ttfamily\bfseries,
    columns = fullflexible,
    aboveskip=1pt,
    belowskip=1pt,
    breaklines = true,
    captionpos = b,
    commentstyle = \color{codeblue},
    keywordstyle = \color{codekw},
}

\definecolor{green}{HTML}{009000}
\definecolor{red}{HTML}{ea4335}

\title{\includegraphics[scale=0.05, bb=0 50 230 0]{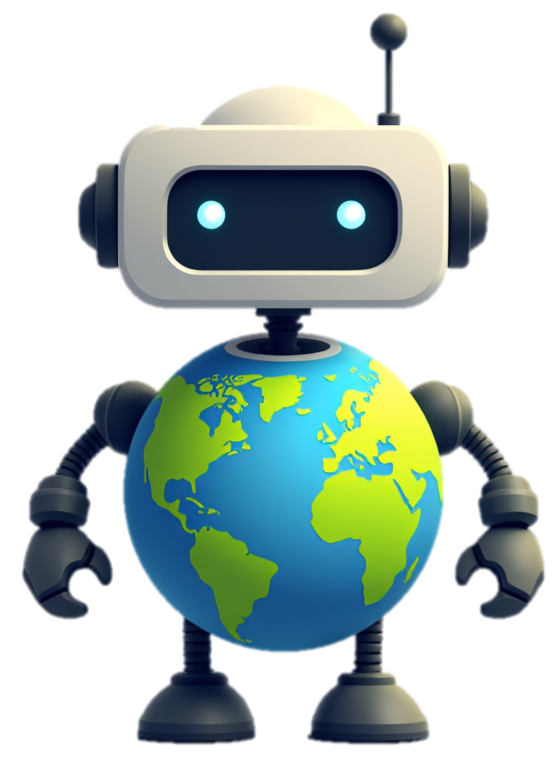} WorldVLA: Towards Autoregressive Action World Model}

\author[1, 2, 3]{Jun Cen}
\author[1]{Chaohui Yu}
\author[1, 2, 3]{Hangjie Yuan}
\author[1]{Yuming Jiang}
\author[1, 2]{Siteng Huang}
\author[1]{Jiayan Guo}
\author[1, 2]{Xin Li}
\author[1]{Yibing Song}
\author[1, 2]{Hao Luo}
\author[1]{Fan Wang}
\author[1, 2]{Deli Zhao}
\author[3]{Hao Chen}

\affiliation[1]{DAMO Academy, Alibaba Group}
\affiliation[2]{Hupan Lab}
\affiliation[3]{Zhejiang University}

\abstract{
We present WorldVLA, an autoregressive action world model that unifies action and image understanding and generation. Our WorldVLA intergrates Vision-Language-Action (VLA) model and world model in one single framework. The world model predicts future images by leveraging both action and image understanding, with the purpose of learning the underlying physics of the environment to improve action generation. Meanwhile, the action model generates the subsequent actions based on image observations, aiding in visual understanding and in turn helps visual generation of the world model. We demonstrate that WorldVLA outperforms standalone action and world models, highlighting the mutual enhancement between the world model and the action model. In addition, we find that the performance of the action model deteriorates when generating sequences of actions in an autoregressive manner. This phenomenon can be attributed to the model's limited generalization capability for action prediction, leading to the propagation of errors from earlier actions to subsequent ones. To address this issue, we propose an attention mask strategy that selectively masks prior actions during the generation of the current action, which shows significant performance improvement in the action chunk generation task.
}

\date{\today} 
\code{https://github.com/alibaba-damo-academy/WorldVLA}
\correspondence{cenjun.cen@alibaba-inc.com}

\begin{document}
\thispagestyle{firstheader}
\maketitle
\pagestyle{empty}

\section{Introduction} \label{sec:introduction}

The development of Vision-Language-Action (VLA) models has emerged as a significant focus within robotics action model research \citep{brohan2023rt, kim2024openvla, black2024pi_0}. These models are constructed by augmenting large-scale pre-trained Multimodal Large Language Models (MLLMs) \citep{liu2023visual, li2024llava, zhang2025videollama, bai2025qwen2} with with either an action head or additional action expert module to generate actions. MLLMs contribute robust capabilities in perception and decision making, enabling VLA models to exhibit enhanced generalization across a wide range of robotic tasks \citep{black2024pi_0, intelligence2025pi_}. Nevertheless, a notable limitation persists: these models often lack a comprehensive understanding of actions, as actions are treated solely as outputs but not being integrated as inputs for deeper analysis.
In contrast, world models demonstrate the ability to predict future visual states based on current observations and actions, thereby achieving a dual understanding of both visual information and behavioral dynamics \citep{ha2018world, agarwal2025cosmos, wu2025ivideogpt}. Despite this advantage, world models are constrained by their inability to directly generate action outputs, resulting in a functional gap that limits their application in scenarios requiring explicit action planning.

\begin{figure}[t!]
    \centering
    \includegraphics[width=0.98\linewidth]{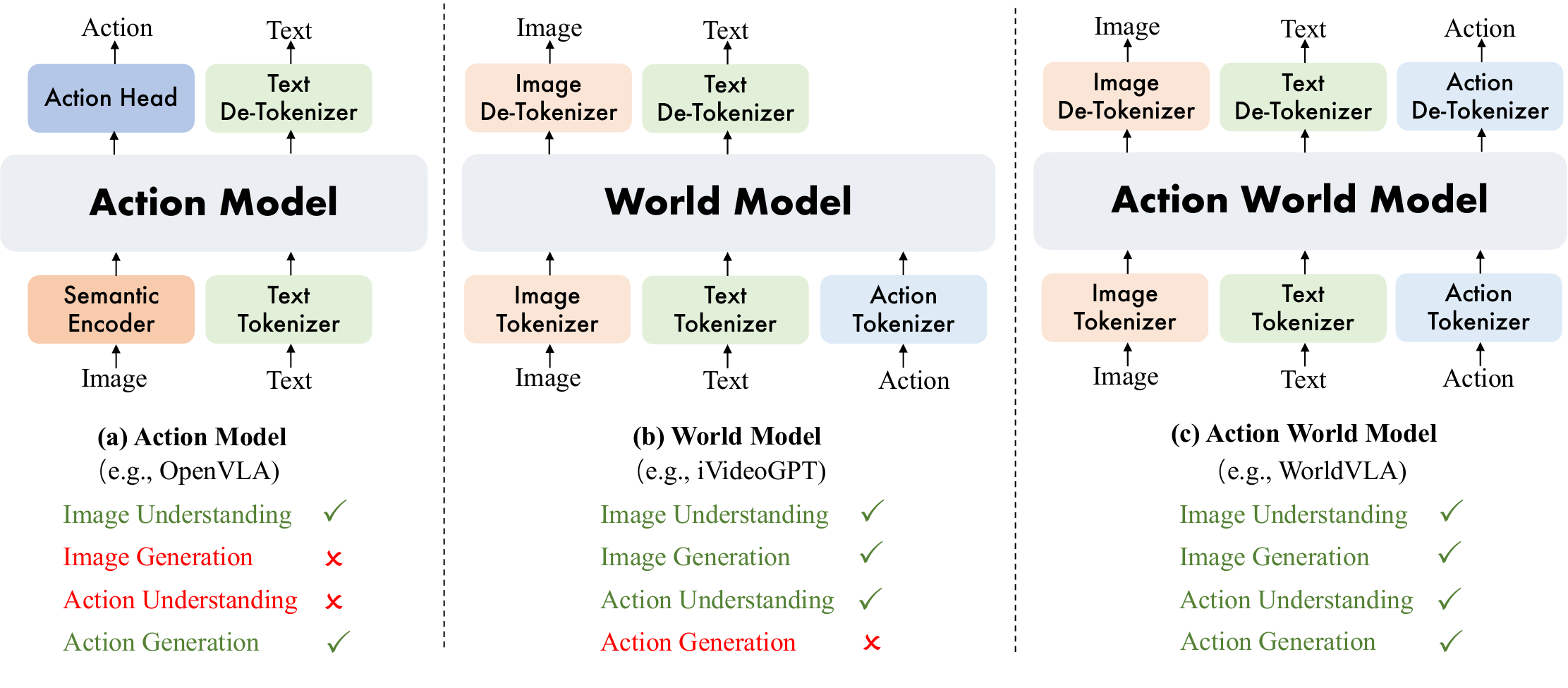}
    \caption{(a) Action model generates actions based on image understanding; (b) World model generates the image based on image and action understanding; (c) Action World Model unifies both image and action understanding and generation.}
    \label{fig:teaser}
\end{figure}

To address the constraints inherent in both Vision-Language-Action (VLA) models and world models, we introduce WorldVLA, an autoregressive action world model for unified action and image understanding and generation. As depicted in Fig.~\ref{fig:teaser}, WorldVLA employs three separate tokenizers to encode images, text, and actions. The tokens from different modalities are set to share the same vocabulary so that understanding and generation across these modalities can be unified within a single LLM architecture. The world model component captures the underlying physical dynamics of the environment by generating visual representations based on input actions. This process of action interpretation and environmental physics learning is essential for enabling effective decision making within the action model. Concurrently, the action model embedded within WorldVLA refines the understanding of visual data, thereby improving the precision of image generation performed by the world model. This bidirectional enhancement creates a more robust and comprehensive model capable of both understanding and generating actions and images.

Action chunking and parallel decoding have been demonstrated to significantly influence the performance of action models~\citep{kim2025fine}. However, we find that generating multiple actions in sequence leads to performance drop in autoregressive models. The primary reason for this is that pretrained multimodal language models have predominantly been exposed to images and text rather than actions, resulting in limited action generalization capabilities. In autoregressive models where subsequent actions are conditioned on preceding ones, error propagation becomes a critical issue, as the earlier incorrect predictions influence subsequent actions over time. To alleviate this issue, we propose an action attention masking strategy that selectively masks prior actions during the generation of current actions. This approach effectively mitigates error accumulation and yields substantial improvements in the task of action chunk generation.

The experiments on LIBERO benchmark show that our WorldVLA outperforms the action model with the same backbone by 4\% grasping success rate. Further, compared to vanilla world model, our WorldVLA shows superior video generation capability and reduces Fréchet Video Distance (FVD) on LIBERO dataset by 10\%. These results underscore the mutual benefits derived from integrating world and action models, highlighting the advantages of a unified framework for image and action comprehension and generation. In the context of action chunk generation, the grasping success rate decreseas by 10\% to 50\% when employing a conventional autoregressive approach. However, the implementation of our attention masking strategy significantly mitigates this decrease, yielding a 4\% to 23\% improvement in grasping success rate.

In summary, our contributions are as follows:
\begin{itemize}
    \item We propose WorldVLA, an autoregressive action world model that unifies action and image understanding and generation.
    \item We introduce an action attention masking strategy for the action chunk generation task in autoregressive models, addressing the challenge of action error accumulation when generating multiple actions in sequence.
    \item Our experiments demonstrate that WorldVLA outperforms the standalone action and world models, highlighting the mutual enhancement between the world model and action model. Additionally, the action attention masking strategy solves the performance degradation when generating action chunks and significantly improves grasping performance.
\end{itemize}


\section{Related Works}
Our proposed WorldVLA is related to the action model, video prediction model, and world model. The difference between them are summaried in Table~\ref{tab:diff_models}.

\begin{table}[t]
\caption{Comparason of different action and video generative models. T: Text; V: Video; A: Action.
}
\centering
\setlength{\tabcolsep}{1.3mm}
\begin{tabular}{lcccccc}
\toprule
{Model Type}
 &Discrete & Continous & Input & Output\\
\midrule
Action Model &OpenVLA~\citep{kim2024openvla} &$\pi0$~\citep{black2024pi_0} & T + V & A  \\
Video Prediction Model &MAGVIT~\citep{yu2023magvit} &SVD~\citep{blattmann2023stable} & T + V & V \\
World Model &iVideoGPT~\citep{wu2025ivideogpt} &DWS~\citep{he2025pre} & T + V + A & V \\
Action World Model &WorldVLA (ours) &UVA~\citep{li2025uva} &T + V + A & V + A \\
\bottomrule
\label{tab:diff_models}
\end{tabular}
\end{table}

\paragraph{Vision-Language-Action Model.}
Behavior cloning~\citep{pomerleau1988alvinn} is a classic imitation learning approach for robot manipulation, which learns a policy by mimicking expert observation-action pairs. Conventional architectures typically combine a vision backbone, such as ResNet~\citep{he2016deep} or Vision Transformer~\citep{dosovitskiy2020image}, with an action head. The action head may consist of multilayer perceptrons (MLPs)~\citep{rumelhart1986learning}, query-based transformer decoders~\citep{zhao2023learning}, or diffusion-based policy heads~\citep{chi2023diffusion}. Recently, Vision-Language-Action (VLA) models have been proposed, utilizing pre-trained multi-modality large language models (MLLM) as the backbone~\citep{brohan2022rt, brohan2023rt, li2023vision, huang2023embodied, belkhale2024minivla, wen2025tinyvla, zhen20243d}. These frameworks are equiped with either discrete action decoders~\citep{kim2024openvla, pertsch2025fast} or continuous diffusion policy heads~\citep{black2024pi_0, wen2024diffusion} to predict actions. The internet-scale prior knowledge in MLLM enables effective generalization to unseen scenarios tasks for VLA models. Our proposed WorldVLA advances this paradigm by jointly generating actions and predicting future video frames, providing a comprehensive solution for understanding and generation.

\paragraph{Video Generation.}
Video generation plays a dual role in robotics. On one hand, some policy models generate the future video first and then generate the corresponding actions based on the generated video~\citep{du2023learning, ajay2023compositional, bu2024closed}. Large-scale video data could be used for pre-training the future video generation part, as seen in approaches~\citep{wu2023unleashing, cheang2024gr}. Here, video generation serves as a mechanism for visual imagination and planning, providing valuable insights that improve downstream policy generation~\citep{cen2024using}. On the other hand, video generation models can act as world models, simulating diverse future scenarios~\citep{ha2018world}. Such world models are widely utilized to generate varied training data~\citep{agarwal2025cosmos}, support model-based reinforcement learning algorithms~\citep{wu2025ivideogpt}, and aid in selecting the most suitable policies from a pool of generated options~\citep{li2025uva, bar2024navigationworldmodels}. In this work, we show that our WorldVLA enables precise control over video generation through action inputs, while also demonstrating that video generation significantly enhances the quality of action generation.

\paragraph{Unified Understanding and Generation Model.}
Most multi-modality large language models (MLLMs) are designed to perform visual understanding tasks, where the model generates textual responses based on combined image and language inputs~\citep{liu2023visual, li2024llava, zhang2025videollama, bai2025qwen2}. Recently, there has been a growing interest in unifying visual understanding and visual generation within a single framework~\citep{team2024chameleon, zhou2024transfusion}. One line of work tokenizes images into discrete tokens akin to text, enabling large language models (LLMs) to both interpret and generate visual content seamlessly~\citep{team2024chameleon, wang2024emu3}. Another approach integrates diffusion processes into LLMs for image generation while relying on additional visual encoders, such as CLIP~\citep{radford2021learning, zhai2023sigmoid}, for image understanding~\citep{chen2025janus, tong2024metamorph}. In the robotics domain, the Unified Video Action Model~\citep{li2025uva} proposes a unified architecture that generates images and actions through distinct diffusion heads. In contrast, our WorldVLA explores an alternative direction by employing a discrete autoregressive architecture to build a unified model capable of handling both perception and action generation.
\section{Methods}

\begin{figure}[t]
    \centering
    \includegraphics[width=0.98\linewidth]{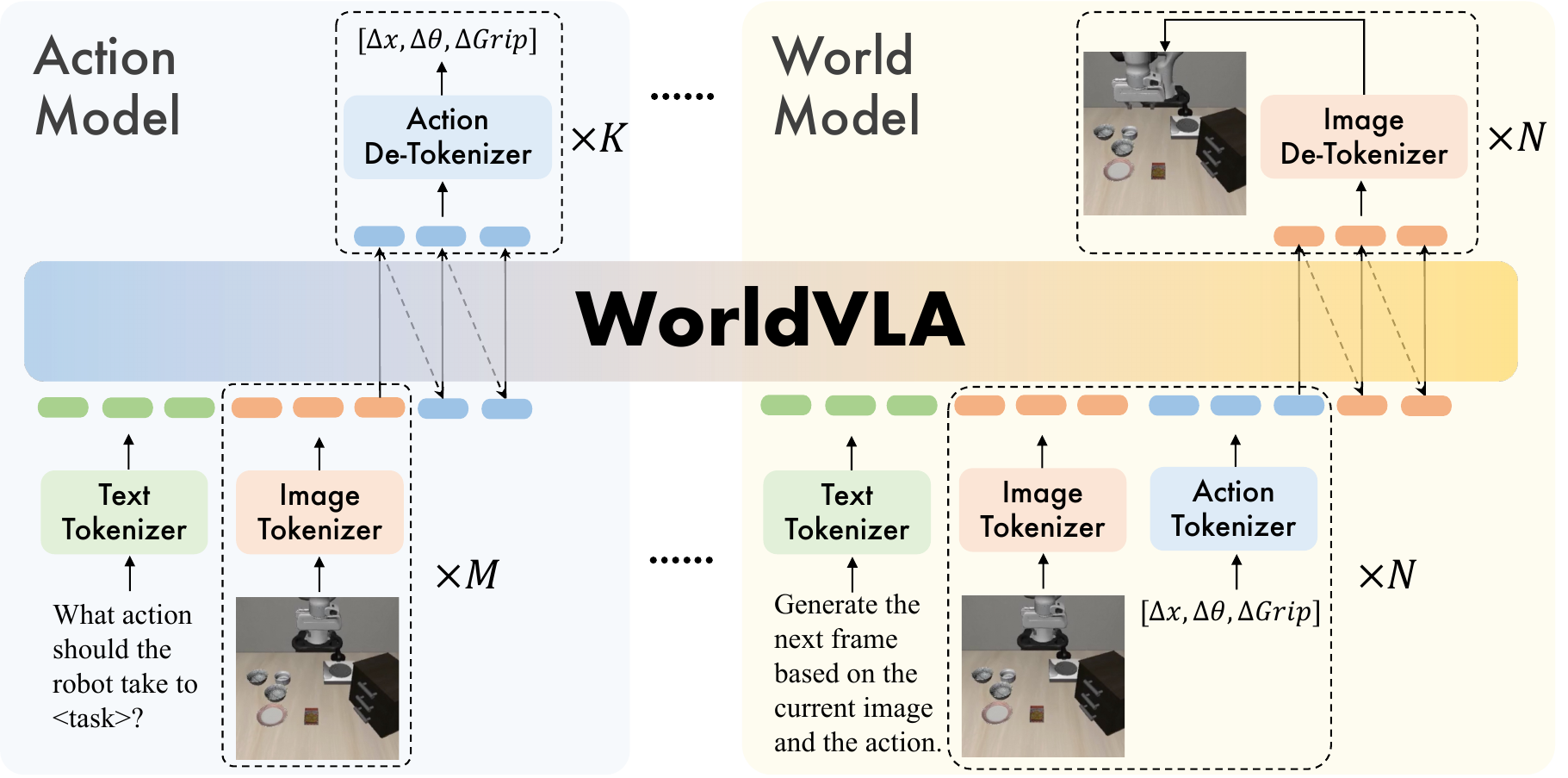}
    \caption{Overview of WorldVLA. WorldVLA integrates two distinct but complementary functional components: an action model and a world model. The action model is responsible for generating actions conditioned on both textual and visual data. The world model functions to predict the subsequent environmental state (e.g., the next visual frame) by leveraging textual information, current image, and current action.}
    \label{fig:outline}
\end{figure}

\subsection{Problem Formulation}

In this work, we address the challenge of learning a unified model capable of simultaneously performing action prediction and world state forecasting. Specifically, we define two primary components: an action model (or policy model) $\pi_\theta$ and a world model $f_\phi$. The action model $\pi_\theta$ is responsible for generating an action $a_t$ conditioned on a history of image observations $\{o_{t-h}, o_{t-h+1}, \dots, o_t\}$ and a language instruction $l$, which can be formally expressed as:
\begin{equation}
a_t = \pi_\theta(a_t \mid o_{t-h:t}, l).
\end{equation}
Meanwhile, the world model $f_\phi$ predicts the next frame $o_t$ based on the historical sequence of observations $\{o_{t-h}, o_{t-h+1}, \dots, o_{t-1}\}$ and the corresponding sequence of actions $\{a_{t-h}, a_{t-h+1}, \dots, a_{t-1}\}$. This relationship is formulated as:
\begin{equation}
o_t = f_\phi(o_t \mid o_{t-h:t-1}, a_{t-h:t-1}).
\end{equation}

Our objective is to develop an integrated action-world model $M_\psi$ that unifies these two functionalities. The model $M_\psi$ should be capable of both predicting actions as a policy model and forecasting future states as a world model. Formally, the unified model $M_\psi$ is defined as:
\begin{equation}
M_\psi: 
\begin{cases}
a_t = M_\psi^{\text{policy}}(a_t \mid o_{t-h:t}, l), \\
o_t = M_\psi^{\text{world}}(o_t \mid o_{t-h:t-1}, a_{t-h:t-1}),
\end{cases}
\end{equation}
where $M_\psi^{\text{policy}}$ represents the action generation component and $M_\psi^{\text{world}}$ denotes the world state prediction component. By learning such a unified model, we aim to achieve a compact and efficient framework that leverages shared representations for both decision-making and environment modeling.

\subsection{Architecture}
The overall architecuture of autoregressive action world model is shown in Fig.~\ref{fig:outline}. We initilize the model from Chameleon~\citep{team2024chameleon} since it is a unified model for image understanding and generation. Three tokenizers are involved, including an image tokenizer, a text tokenizer, and an action tokenizer. The image tokenizer is a VQ-GAN model~\citep{esser2021taming} with additional perceptual losses to specific image regions, \textit{e.g.}, faces and salient objects~\citep{gafni2022make}. The compression ratio of the image tokenizer is 16 and the codebook size is 8192. The image tokenizer generates 256 tokens for $256 \times 256$ images and 1024 tokens for $512 \times 512$ images. The action tokenizer discretizes each dimension of continuous robot actions into one of 256 bins, with bin widths determined by the range of the training data~\citep{kim2024openvla, brohan2023rt}. The actions are represented as 7 tokens, including 3 relative positions, 3 relative angles and 1 absolute gripper states. The text tokenizer is a trained BPE tokenizer~\citep{sennrich2015neural} with a vocabulary size of 65,536, which includes 8192 image tokens and 256 action tokens. All texts, actions, and images are discretized into tokens and are trained under the autoregressive manner.

\begin{figure}[t]
    \centering
    \includegraphics[width=0.98\linewidth]{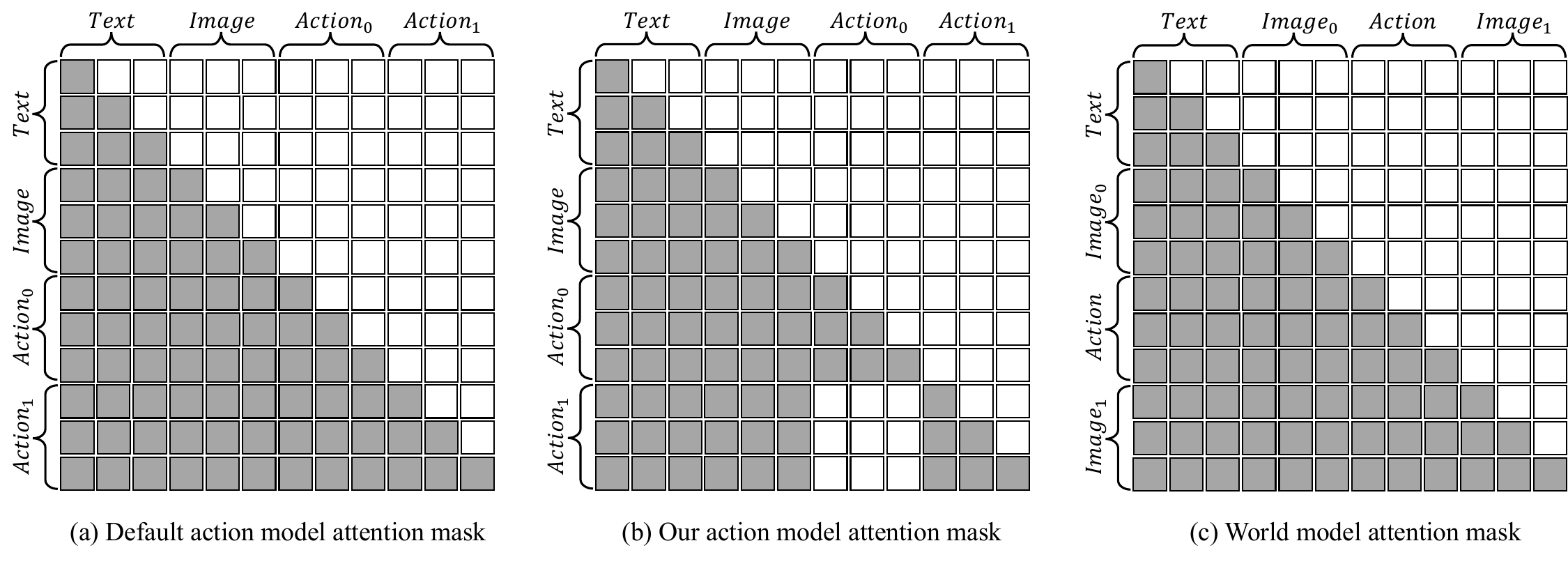}
    \caption{Attention mask mechanism of (a) default action model, (b) our proposed action model, and (c) world model.}
    \label{fig:attention_mask}
\end{figure}

\subsection{Training Strategy}

We mix the action model data and world model data to train our WorldVLA. There are three primary reasons for incorporating world model data to enhance action generation. First, the world model acquires an understanding of the environmental physics by learning to predict future observations based on the current state and applied actions. This learned representation of environmental physics is helpful for manipulation tasks. Second, the world model enables the system to simulate and evaluate potential outcomes of candidate actions, thereby facilitating the avoidance of actions that may lead to unfavorable states. Third, the world model requires a precise interpretation of the action inputs, which in turn supports the action model in producing more effective and contextually appropriate actions. On the other hand, action model enhances visual understanding and in turn supports the visual generation capability of the world model.

\paragraph{Action Model Data.}
Action model is to generate the action given the text instruction and image observations. The text inputs are \textit{"What action should the robot take to + task instruction + ?"}. The overall token sequence is:

\begin{center}
\texttt{[BOS]\{text\}}$\underbrace{\texttt{[BOI]\{image\}\dots\{image\}[EOI]}}_{\times M}$\texttt{[EOS]}$\overbrace{\underbrace{\texttt{[BOA]\{action\}\dots\{action\}[EOA]}}_{\times K}\texttt{[EOS]}}^{\mathcal{L}_{action}}$,
\end{center}

where \texttt{\{text\}}, \texttt{\{image\}}, and \texttt{\{action\}} refer to the discret text, image, and action tokens. \texttt{[BOS]}, \texttt{[EOS]}, \texttt{[BOI]}, \texttt{[EOI]}, \texttt{[BOA]}, \texttt{[EOA]} refer to the beginning of sentence, end of sentence, beginning of image, end of image tokens, beginning of action, and end of action tokens. The input contains $M$ images and the output contains $K$ actions. We only calculate the loss of action tokens $\mathcal{L}_{action}$.

\begin{table}[t!]
\centering
\setlength{\tabcolsep}{3.2mm}
\caption{Evaluation results on LIBERO benchmark. Pretraining means the model is pretrained on the large-scale robot manipulation data.}
\begin{tabular}{lcccccc}
\toprule
\textbf{Continous Action Model} & Pretraining & \begin{tabular}[c]{@{}c@{}}Spatial \end{tabular} & \begin{tabular}[c]{@{}c@{}}Object \end{tabular} & \begin{tabular}[c]{@{}c@{}}Goal \end{tabular} & \begin{tabular}[c]{@{}c@{}}Long \end{tabular} & \begin{tabular}[c]{@{}c@{}}Average \end{tabular} \\ \midrule
Diffusion Policy~\citep{chi2023diffusion} &\ding{55}                            & 78.3 & 92.5 & 68.3 & 50.5 & 72.4 \\
Octo~\citep{team2024octo} &\ding{51}                                           & 78.9 & 85.7 & 84.6 & 51.1 & 75.1 \\
DiT Policy~\citep{hou2024diffusion} &\ding{51}                                   & 84.2 & 96.3 & 85.4 & 63.8 & 82.4 \\
Seer~\citep{tian2024predictive} &\ding{55}                                              & -- & -- & -- & 78.7 & -- \\
Seer~\citep{tian2024predictive} &\ding{51}  & -- & -- & -- & 87.7 & -- \\ 
OpenVLA-OFT~\citep{kim2025fine} &\ding{51} & {96.9} & 98.1 & 95.5 & {91.1} & {95.4} \\
UVA~\citep{li2025uva}&\ding{55} & --& --& --& 93.0& -- \\
\midrule
\textbf{Discrete Action Model} \\
\midrule
OpenVLA~\citep{kim2024openvla} &\ding{51}                                       & 84.7 & 88.4 & 79.2 & 53.7 & 76.5 \\
WorldVLA (256 $*$ 256) &\ding{55} &85.6 &89.0 &82.6 &59.0 &79.1\\
WorldVLA (512 $*$ 512) &\ding{55} &87.6 &96.2 &83.4 &60.0 &81.8\\
\bottomrule
\end{tabular}
\label{tab:bench}
\end{table}

\paragraph{World Model Data.} World model is to generate the next image frame given the current image observation and action. It does not need the task instruction since the action itself could totally determine the next state. The text inputs are \textit{"Generate the next frame based on the current image and the action."}. The overall token sequence is:

\begin{center}
\texttt{[BOS]\{text\}}
$\underbrace{ 
  \texttt{[BOI]\{image\}[EOI][BOA]\{action\}[EOA][EOS]} 
    \overbrace{\texttt{[BOI]\{image\}[EOI][EOS]} 
}^{\mathcal{L}_{world}}}_{\times N}$ 
. 
\end{center}

The next frame prediction conditioned on the action repeats $N$ times, and we only calculate the loss of generated image tokens $\mathcal{L}_{world}$.

\paragraph{Attention Mask.} The standard attention mechanism in autoregressive models typically employs a causal attention mask, which restricts the current token's access to information exclusively from preceding tokens, excluding any subsequent ones, as illustrated in Fig.~\ref{fig:attention_mask} (a). Nevertheless, this conventional configuration proves inadequate for generating action chunks, \textit{i.e.}, multiple consecutive actions. While the foundational MLLM demonstrates robust generalization capabilities across image and text domains due to the large-scale pretraining on diverse datasets, its capacity to generalize effectively in the action domain is comparatively limited. Consequently, errors originating from earlier actions propagate to subsequent actions under the default attention mask, resulting in performance degradation. To address this limitation, we introduce an alternative attention mask tailored for action generation, depicted in Fig.~\ref{fig:attention_mask} (b). This modified mask ensures that current actions rely solely on textual and visual inputs, while prohibiting access to prior actions. Such a design enables the autoregressive framework to generate multiple actions in parallel, aligning with methodologies presented in~\citep{kim2025fine, black2024pi_0}. The world model part adheres to the conventional causal attention mask, as shown in Fig.~\ref{fig:attention_mask} (c).

\paragraph{Training Objective.}
We mix the action model data and world model data so that the autoregressive action world model could behave as both action model and world model. The loss function is:
\begin{equation}
    \mathcal{L} = \mathcal{L}_{action} + \alpha \mathcal{L}_{world},
\end{equation}
where $\mathcal{L}_{action}$ and $\mathcal{L}_{world}$ refer to the cross-entropy loss of the action model data and world model data. Since the image tokens (256 tokens for $256 \times 256$ images and 1024 tokens for $512 \times 512$ images) are much more than the action tokens (7 tokens), we use $\alpha$ to balance the loss contribution.

\section{Experiments}

\subsection{Evaluation Benchmark}
\paragraph{Benchmark.} We use LIBERO benchmark~\citep{liu2023libero} in our experiments. LIBERO benchmark contains LIBERO-Spatial, LIBERO-Object, LIBERO-Goal, LIBERO-Long and LIBERO-90. LIBERO-Spatial focuses on spatial relationships by requiring the robot to place a bowl based on its location. LIBERO-Object emphasizes object recognition by having the robot pick and place unique objects. LIBERO-Goal tests procedural learning through varying task goals with fixed objects. LIBERO-Long includes 10 long-horizon tasks. LIBERO-90 provides 90 short-horizon tasks for pretraining. 

\paragraph{Datasets.} We first filter out the unsuccessful recorded trajectories and no-operation actions like OpenVLA~\citep{kim2024openvla}. Considering the world model evaluation needs ground truth-paired video and action data, we split the 90\% of the trajectories as the training set and the 10\% remaining trajectories as the validation set. The training set is used for model training by default, with the exception of Table~\ref{tab:bench}, where all available data are utilized during training to ensure a fair comparison.

\paragraph{Baselines.} There are two kinds of action models including the continous action model and discrete action model. Continous action model generates multiple actions in parellel and uses l1 regression loss for training. Diffusion-based action model like Diffusion Policy~\citep{chi2023diffusion}, Octo~\citep{team2024octo}, DiT Policy~\citep{hou2024diffusion}, and UVA~\citep{li2025uva} use diffusion process to generate the actions. Seer~\citep{tian2024predictive} and OpenVLA-OFT~\citep{kim2025fine} use an action head to directly output multiple actions in one time. Discrete action models like OpenVLA~\citep{kim2024openvla} considers the action as tokens just like texts, and the actions are generated in an autoregressive manner. Discrete models inherently exhibit inferior performance, as the tokenization process of actions may lead to information loss.

\paragraph{Training Setting.} The action model utilizes a default input image count of $M=2$. The action chunk size is set to $K=10$ for the LIBERO Long task and $K=5$ for the remaining three LIBERO tasks under default configuration. To minimize computational expenditure, the world model operates with a single round $N=1$. The parameter $\alpha$ is fixed at 0.04 in the experimental setup.

\paragraph{Metrics.} For action model evaluation, each task is evaluated for 50 rollouts under different initial states and we record the success rates (SR). For world model evaluation, we use the validation set and record the FVD, PSNR, SSIM, and LPIPS values. 

\subsection{Evaluation Results and Discussion}

\paragraph{Benchmark Results.}
Table~\ref{tab:bench} indicates that the proposed WorldVLA model exhibits superior performance compared to the discrete OpenVLA model, even in the absence of pretraining. This outcome shows the effectiveness of the WorldVLA's design.
Furthermore, a positive correlation is observed between image resolution and model performance. Specifically, the 512 $*$ 512 pixel resolution yielded enhanced results compared to the 256 $*$ 256 pixel resolution. This phenomenon is primarily attributable to the pretraining regimen of the Chameleon backbone~\citep{team2024chameleon}, whose image tokenization module and the large language model components are inherently optimized at 512 $*$ 512 resolution. Additionally, higher resolution naturally provides a greater level of detailed visual information, which is particularly crucial for robotic grasping tasks since it demands high operational precision.

\begin{table}[t!]
\caption{Ablation study of action model.
}
\centering
\setlength{\tabcolsep}{1.8 mm}
\begin{tabular}{ccccccccccc}
\toprule
Index&
 \makecell{Action \\ Model} & \makecell{World \\ Model} & \makecell{Action \\ Chunking} & \makecell{Our Action Model\\Attention Mask} & \makecell{Goal \\ SR (\%)} & \makecell{Object \\ SR (\%)} & \makecell{Spatial \\ SR (\%)} & \makecell{Long \\ SR (\%)} & \makecell{Average \\ SR (\%)}\\
\midrule
 1& \ding{51} &\ding{55} &\ding{55} &\ding{55}    & 67.3          & 82.9           & 77.8            & 23.0         &62.8\\
2& \ding{51} &\ding{51} &\ding{55} &\ding{55} & 73.1          & 88.0           & 80.2            & 27.3         &67.2          \\
3& \ding{51} &\ding{55} &\ding{51} &\ding{55} &79.6   &82.9 &36.7 &16.9 &54.0      \\
4& \ding{51} &\ding{55} &\ding{51} &\ding{51} & 84.4 &90.9 &81.8 &49.3  &76.6      \\
5& \ding{51} &\ding{51} &\ding{51} &\ding{51} & 85.1 &90.9 &84.0 &52.4 &78.1     \\
\bottomrule
\label{tab:ab_action}
\end{tabular}
\end{table}

\begin{figure}[t!]
    \centering
    \includegraphics[width=0.98\linewidth]{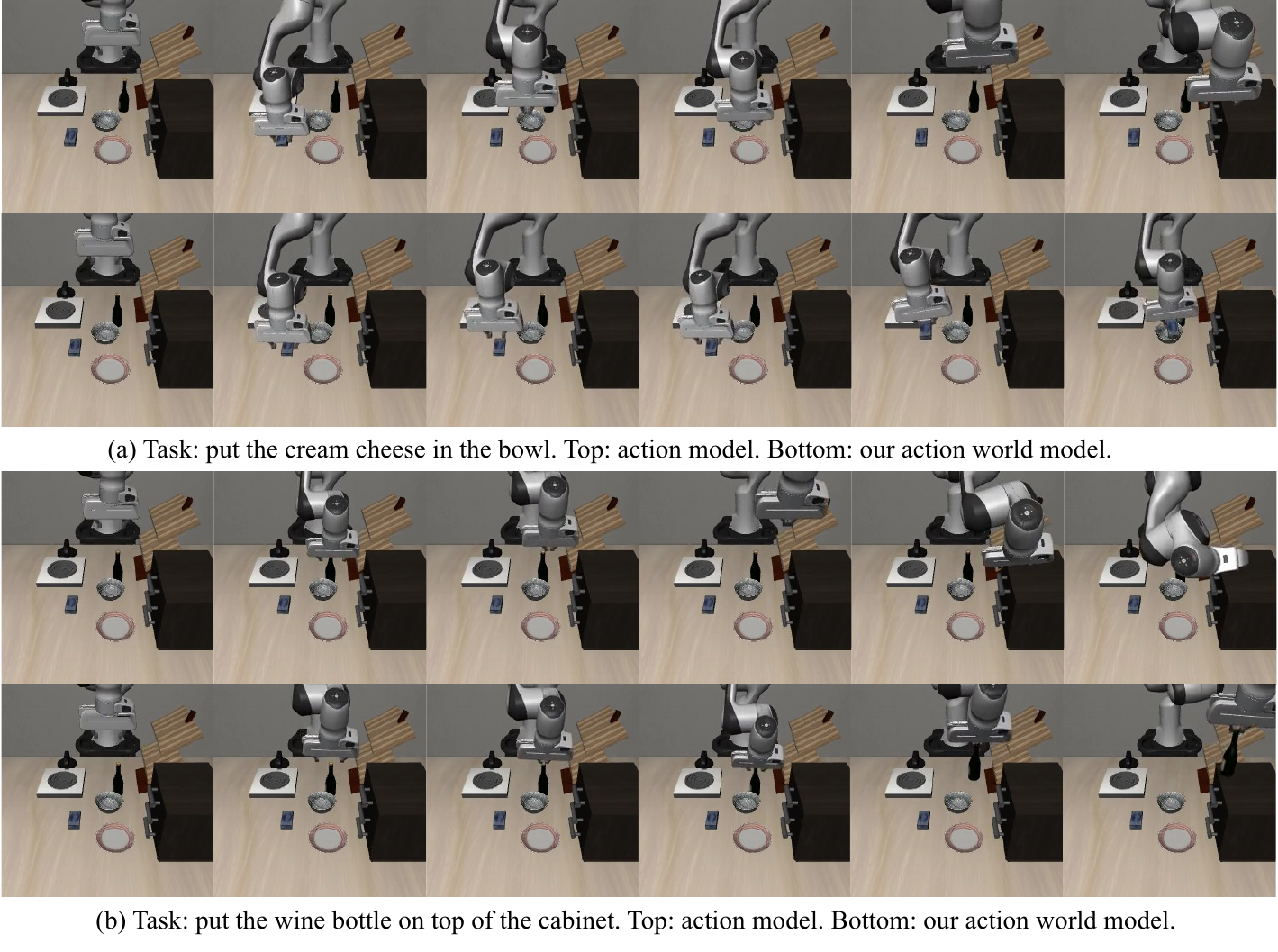}
    \caption{Visualization of action model. Top: action model. Bottom: our action world model.}
    \label{fig:vis_action_model}
\end{figure}

\paragraph{World Model Helps Action Model.} Quantitative results in Table~\ref{tab:ab_action}, including row 2 vs. row 1, or row 5 vs. row 4, demonstrate that the integration of a world model significantly enhances the performance of the action model. The world model's fundamental function involves predicting the subsequent state of the environment, conditioned on the current state and a given action. This generative process inherently promotes the acquisition of an understanding of the system's underlying physical dynamics, which is a critical prerequisite for successful execution in dexterous manipulation tasks such as grasping. Furthermore, the world model endows the system with the capacity for prospective simulation, enabling it to anticipate the consequences of potential actions. This predictive foresight facilitates more informed decision-making, thereby optimizing action selection to maximize the probability of task success. Fig.~\ref{fig:vis_action_model} shows that the action model directly move to the destination without successfully grasping the cheese or bottle. In contrast, our action world model repeatedly attempts to grasp the objects until successful manipulation is achieved before proceeding to the target location.


\begin{figure}[h!]
    \centering
    \includegraphics[width=0.98\linewidth]{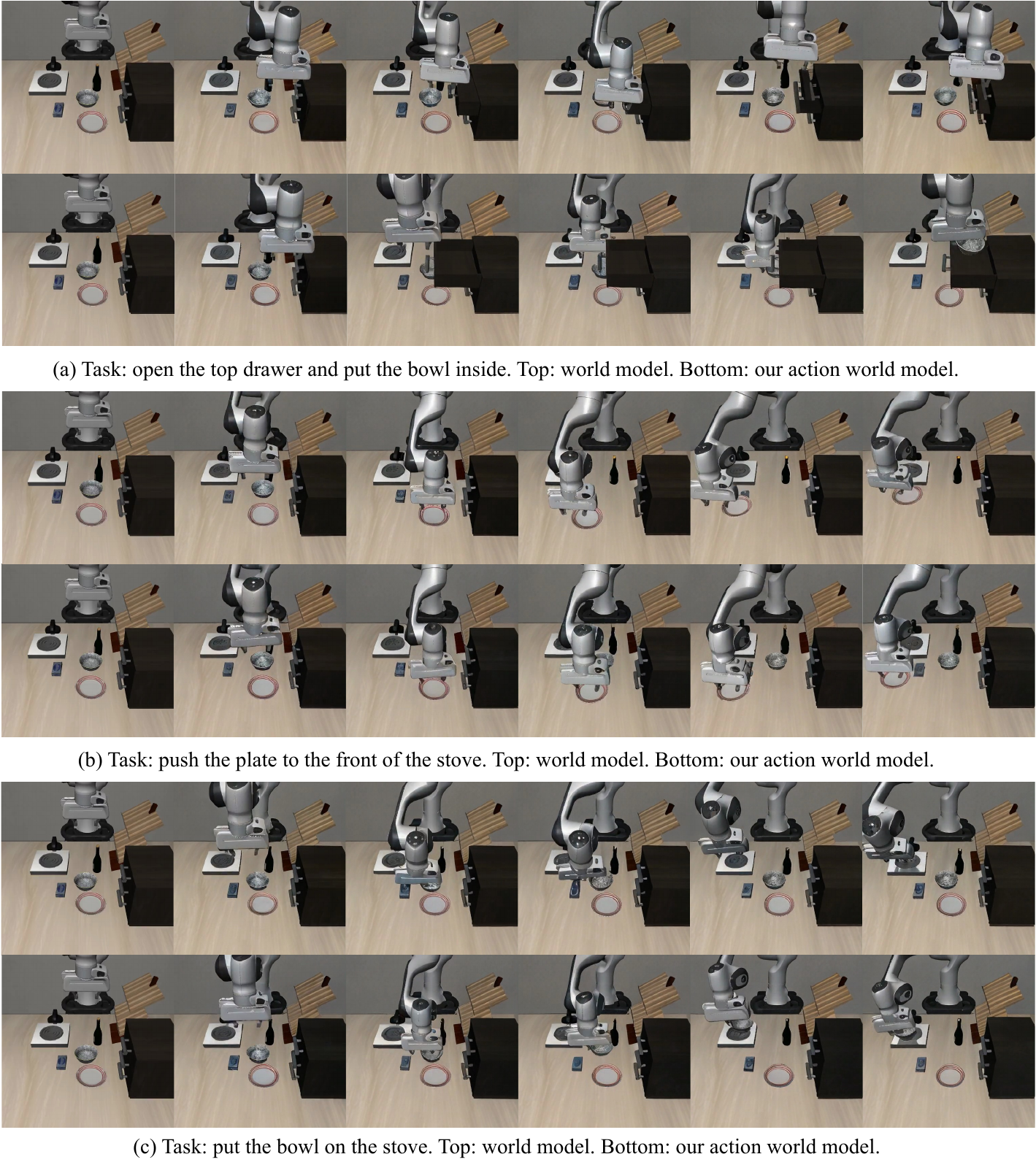}
    \caption{Visualization of world model. Top: world model. Bottom: our action world model.}
    \label{fig:vis_world_model}
\end{figure}

\paragraph{Action Model Helps World Model.} Table~\ref{tab:ab_world} demonstrates that the action world model outperforms the pure world model in terms of generation quality, particularly when producing longer video sequences. The action model derives actions based on the input images. On one hand, this contributes to more accurate visual interpretation; on the other, the process of generating actions enhances the understanding of the underlying behavioral patterns. Both aspects support the overall performance of the world model, which relies on robust comprehension of both visual and action-related information to predict future states effectively. As illustrated in Fig.~\ref{fig:vis_world_model}, the pure world model fails in several scenarios: it is unable to open the drawer (a), causes the bowl to disappear after moving the disk (b), and fails to lift the bowl onto the stove (c). In contrast, the action world model produces coherent and physically plausible subsequent states in these cases.

\begin{table}[t!]
\caption{Ablation study of world model.}
\centering
\setlength{\tabcolsep}{2.6 mm}
\begin{tabular}{lcccccccccc}
\toprule
 & \multicolumn{4}{c}{10 frames} & \multicolumn{4}{c}{50 frames} \\
\cmidrule(lr){2-5} \cmidrule(lr){6-9}
 & {FVD}$\downarrow$ & {PSNR}$\uparrow$ & {SSIM}$\uparrow$ & {LPIPS}$\downarrow$ & {FVD}$\downarrow$ & {PSNR}$\uparrow$ & {SSIM}$\uparrow$ & {LPIPS}$\downarrow$\\
\midrule
World Model &\textbf{250.0} &29.62 &\textbf{90.73} &11.97  &718.6 &23.98 &83.41 &15.60  \\
Action World Model &255.1 &\textbf{29.77} &90.40 &\textbf{11.94} &\textbf{674.1} &\textbf{24.30} &\textbf{83.55} &\textbf{15.44}  \\
\bottomrule
\label{tab:ab_world}
\end{tabular}
\end{table}

\begin{figure}[t!]
    \centering
    \includegraphics[width=0.98\linewidth]{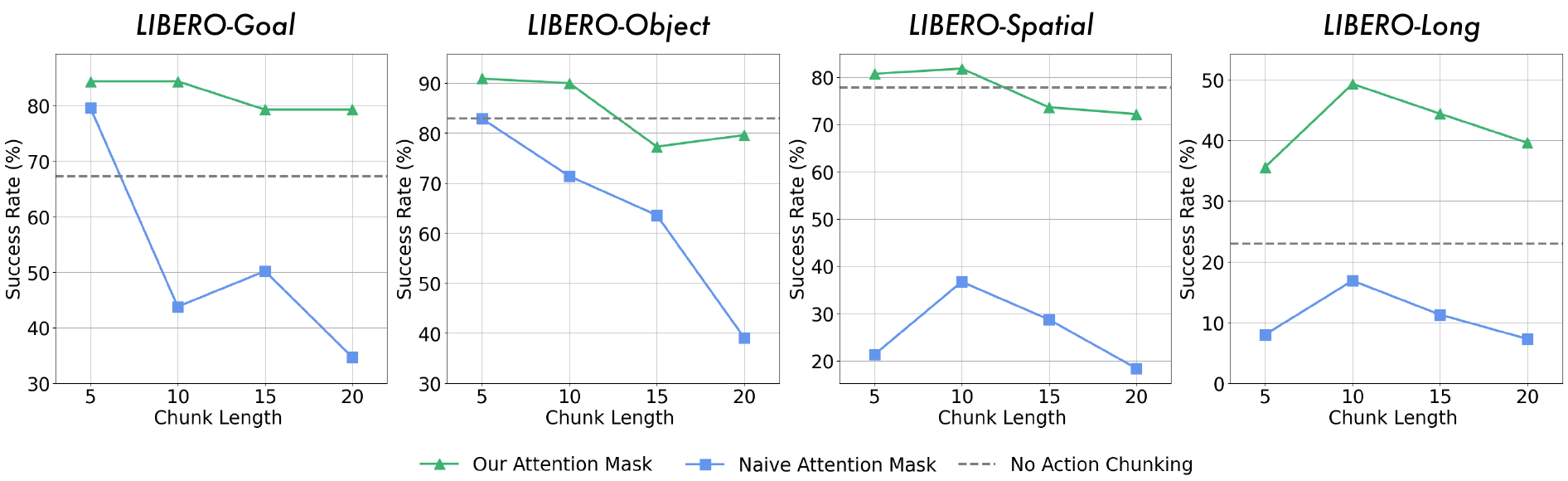}
    \caption{Ablation study of action chucking length.}
    \label{fig:action_ck}
\end{figure}

\paragraph{Action Chunking Generation with Proposed Attention Mask.} Simultaneous generation of multiple actions is essential for achieving effective and efficient grasping. However, we observe that a naive autoregressive approach—where actions are generated sequentially—can degrade model performance, as evidenced by the results in row 3 of Table~\ref{tab:ab_action} and Fig.~\ref{fig:action_ck}. The grasping success rate gradullay decreases with longer action chunks. This degradation arises because later actions become overly dependent on preceding ones since they share the same space, rather than being grounded in visual input which is a distinct modality. The generalization of the action is not that strong as this modality was not involved during pretraining the MLLM. Consequently, errors tend to accumulate as the sequence of generated actions increases. The proposed attention masking mechanism ensures that each action is generated independently and solely determined by the visual input, thereby mitigating the issue of error propagation within the action sequence. As illustrated in Fig.~\ref{fig:action_ck}, the model incorporating the proposed attention mask demonstrates superior performance compared to the naive attention mask, particularly under conditions of longer chunk lengths. This highlights the efficacy of the introduced masking approach. If the length of the action chunk is excessively prolonged, the robot's ability to timely adapt its policy becomes constrained, leading to a decline in overall performance, as demonstrated in Fig.~\ref{fig:action_ck}.

\begin{figure}[t!]
    \centering
    \includegraphics[width=0.98\linewidth]{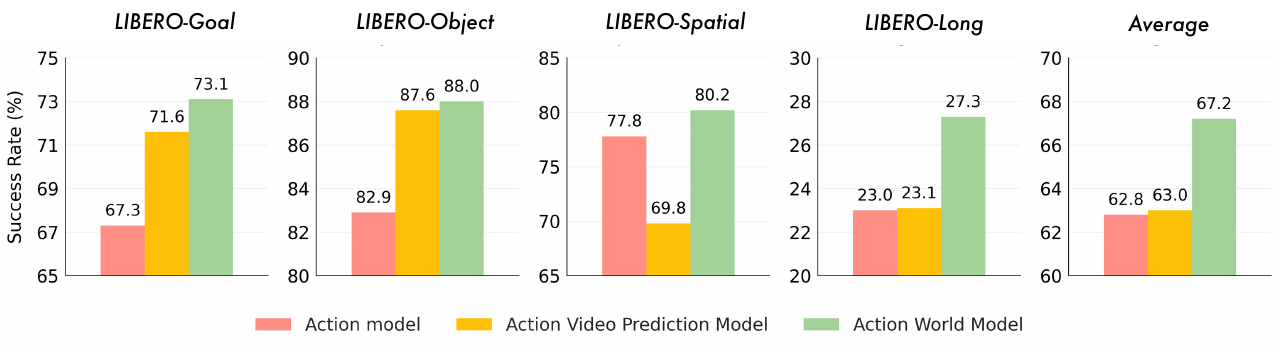}
    \caption{Comparison between action world model and action video prediction model.}
    \label{fig:video_pred}
\end{figure}

\paragraph{World Model versus Video Prediction Model.} Video prediction model is to generate the next frames based on the current frame and the task instruction. Video prediction has been used for pretraining the action model in prior research, such as GR-1~\citep{wu2023unleashing} and GR-2~\citep{cheang2024gr}. Both video prediction model and world model belong to visual generation models, so we conduct a comparison to assess which framework provides greater utility for the action model. The text inputs of video prediction model are \textit{"Generate the future image based on the task and current image. + task instruction"}. The overall token sequence is:
\begin{center}
\texttt{[BOS]\{text\}}$\underbrace{\texttt{[BOI]\{image\}\dots\{image\}[EOI]}}_{\times M}$\texttt{[EOS]}$\overbrace{\underbrace{\texttt{[BOI]\{image\}[EOI]}}_{\times K}\texttt{[EOS]}}^{\mathcal{L}_{video}}$.
\end{center}
The difference between video prediction model and world model is that the world model is conditioned on the action while video prediction model is not. As illustrated in Fig.~\ref{fig:video_pred}, the integration of a world model enhances the performance of the action model across all evaluated tasks. The video prediction model, however, demonstrates beneficial effects for two tasks while negatively impacting performance on one task. This discrepancy may arise from the inherent ambiguity in video prediction when action inputs are absent, as the subsequent frame cannot be uniquely determined from the initial frame alone. Consequently, multiple plausible future frames or ground truth sequences may correspond to a single starting frame, potentially introducing noise or inconsistency during training. Furthermore, the incorporation of a world model necessitates an understanding of actions which could contribute to more effective action generation.

\begin{table}[t!]
\caption{Ablation study of historical image input length.}
\centering
\setlength{\tabcolsep}{4.5 mm}
\begin{tabular}{lcccccccccc}
\toprule
 & \multicolumn{2}{c}{1 frame} & \multicolumn{2}{c}{2 frames} & \multicolumn{2}{c}{4 frames} \\
\cmidrule(lr){2-3} \cmidrule(lr){4-5} \cmidrule(lr){6-7}
 & {SR (\%)}$\uparrow$ & {FPS}$\uparrow$ & {SR (\%)}$\uparrow$ & {FPS}$\uparrow$ & {SR (\%)}$\uparrow$ & {FPS}$\uparrow$\\
\midrule
w/o Action Chunking &{58.4} & 2.27
&67.3  &1.77 &78.7 &1.22\\
w/ Action Chunking  &74.0 &{3.67} &84.4 &{3.13} &{84.7} &{2.78}  \\
\bottomrule
\label{tab:his_input}
\end{tabular}
\end{table}

\paragraph{Historical Image Input.} Unified models for understanding and generation, such as Chameleon~\citep{team2024chameleon}, employ the discrete image tokenizer VQGAN~\citep{esser2021taming} for image interpretation. However, their capacity for semantic comprehension is comparatively limited when contrasted with vision-based perceptual models like CLIP~\citep{radford2021learning}. As demonstrated in Table~\ref{tab:his_input}, the use of a single-frame input results in suboptimal performance. To enhance the model's access to visual context, we incorporate multiple historical image frames, which leads to a progressive improvement in performance. Furthermore, the results indicate that the performance is saturated with two frames when generating action chunks. Consequently, we adopt a two-frame input configuration as the default in our experiments, optimizing the trade-off between task success rate and computational efficiency.


\begin{table}[t!]
\caption{Ablation study of world model pretraining.
}
\centering
\setlength{\tabcolsep}{0.8 mm}
\begin{tabular}{lcccccc}
\toprule
 & {Goal SR (\%)} & {Object SR (\%)} & {Spatial SR (\%)} & {Long SR (\%)} & {Average SR (\%)}\\
\midrule
w/o World Model Pretrain    & 67.3          & 82.9           & 77.8            & 23.0         &62.8\\
w/ World Model Pretrain      & \textbf{73.1}  &\textbf{84.0} &\textbf{79.8} &\textbf{30.2}  &\textbf{66.8}     \\
\bottomrule
\label{tab:pretrain}
\end{tabular}
\end{table}

\paragraph{Pretrain Action Model using World Model.} Our WorldVLA framework integrates both action model data and world model data during training. We further investigate the possibility of utilizing the world model as a source of pretraining weights for the action model. This form of pretraining necessitates that the model develop an understanding of visual inputs, actions, and the underlying physical dynamics governing state transitions. As presented in Table~\ref{tab:pretrain}, employing the world model for pretraining leads to notable improvements in grasping performance. These findings highlight the potential of leveraging world model pretraining in robotic applications, particularly in enhancing task-specific performance through prior exposure to general world knowledge.
\section{Conclusion and Future Work}

This study introduces WorldVLA, a novel autoregressive framework that unifies action and visual understanding with generation capabilities. We demonstrate that the integration of world modeling and action modeling within this architecture can lead to mutual enhancement in performance. An attention mask mechanism has been proposed to enable autoregressive generation of action sequences.
Scaling of both data and model size emerges as a promising avenue for further development of the WorldVLA framework. Additionally, the current image tokenizer, which relies on discrete representations, exhibits limitations in perceptual expressiveness; hence, the design of a unified tokenizer capable of both understanding and generating high-quality visual content is an important direction for improvement. The incorporation of an auxiliary action head presents another potential strategy to enhance grasping performance.
We anticipate that this work will contribute to and inspire future research in robotics, particularly in the domains of world modeling and unified models for action and image understanding and generation.

\bibliographystyle{assets/plainnat}
\bibliography{paper}

\begin{thebibliography}{51}
\providecommand{\natexlab}[1]{#1}
\providecommand{\url}[1]{\texttt{#1}}
\expandafter\ifx\csname urlstyle\endcsname\relax
  \providecommand{\doi}[1]{doi: #1}\else
  \providecommand{\doi}{doi: \begingroup \urlstyle{rm}\Url}\fi

\bibitem[Agarwal et~al.(2025)Agarwal, Ali, Bala, Balaji, Barker, Cai, Chattopadhyay, Chen, Cui, Ding, et~al.]{agarwal2025cosmos}
Niket Agarwal, Arslan Ali, Maciej Bala, Yogesh Balaji, Erik Barker, Tiffany Cai, Prithvijit Chattopadhyay, Yongxin Chen, Yin Cui, Yifan Ding, et~al.
\newblock Cosmos world foundation model platform for physical ai.
\newblock \emph{arXiv preprint arXiv:2501.03575}, 2025.

\bibitem[Ajay et~al.(2023)Ajay, Han, Du, Li, Gupta, Jaakkola, Tenenbaum, Kaelbling, Srivastava, and Agrawal]{ajay2023compositional}
Anurag Ajay, Seungwook Han, Yilun Du, Shuang Li, Abhi Gupta, Tommi Jaakkola, Josh Tenenbaum, Leslie Kaelbling, Akash Srivastava, and Pulkit Agrawal.
\newblock Compositional foundation models for hierarchical planning.
\newblock \emph{Advances in Neural Information Processing Systems}, 36:\penalty0 22304--22325, 2023.

\bibitem[Bai et~al.(2025)Bai, Chen, Liu, Wang, Ge, Song, Dang, Wang, Wang, Tang, et~al.]{bai2025qwen2}
Shuai Bai, Keqin Chen, Xuejing Liu, Jialin Wang, Wenbin Ge, Sibo Song, Kai Dang, Peng Wang, Shijie Wang, Jun Tang, et~al.
\newblock Qwen2. 5-vl technical report.
\newblock \emph{arXiv preprint arXiv:2502.13923}, 2025.

\bibitem[Bar et~al.(2024)Bar, Zhou, Tran, Darrell, and LeCun]{bar2024navigationworldmodels}
Amir Bar, Gaoyue Zhou, Danny Tran, Trevor Darrell, and Yann LeCun.
\newblock Navigation world models, 2024.
\newblock \url{https://arxiv.org/abs/2412.03572}.

\bibitem[Belkhale and Sadigh(2024)]{belkhale2024minivla}
Suneel Belkhale and Dorsa Sadigh.
\newblock Minivla: A better vla with a smaller footprint, 2024.
\newblock \url{https://github.com/Stanford-ILIAD/openvla-mini}.

\bibitem[Black et~al.(2024)Black, Brown, Driess, Esmail, Equi, Finn, Fusai, Groom, Hausman, Ichter, et~al.]{black2024pi_0}
Kevin Black, Noah Brown, Danny Driess, Adnan Esmail, Michael Equi, Chelsea Finn, Niccolo Fusai, Lachy Groom, Karol Hausman, Brian Ichter, et~al.
\newblock A vision-language-action flow model for general robot control.
\newblock \emph{arXiv preprint arXiv:2410.24164}, 2024.

\bibitem[Blattmann et~al.(2023)Blattmann, Dockhorn, Kulal, Mendelevitch, Kilian, Lorenz, Levi, English, Voleti, Letts, et~al.]{blattmann2023stable}
Andreas Blattmann, Tim Dockhorn, Sumith Kulal, Daniel Mendelevitch, Maciej Kilian, Dominik Lorenz, Yam Levi, Zion English, Vikram Voleti, Adam Letts, et~al.
\newblock Stable video diffusion: Scaling latent video diffusion models to large datasets.
\newblock \emph{arXiv preprint arXiv:2311.15127}, 2023.

\bibitem[Brohan et~al.(2022)Brohan, Brown, Carbajal, Chebotar, Dabis, Finn, Gopalakrishnan, Hausman, Herzog, Hsu, et~al.]{brohan2022rt}
Anthony Brohan, Noah Brown, Justice Carbajal, Yevgen Chebotar, Joseph Dabis, Chelsea Finn, Keerthana Gopalakrishnan, Karol Hausman, Alex Herzog, Jasmine Hsu, et~al.
\newblock Rt-1: Robotics transformer for real-world control at scale.
\newblock \emph{arXiv preprint arXiv:2212.06817}, 2022.

\bibitem[Brohan et~al.(2023)Brohan, Brown, Carbajal, Chebotar, Chen, Choromanski, Ding, Driess, Dubey, Finn, et~al.]{brohan2023rt}
Anthony Brohan, Noah Brown, Justice Carbajal, Yevgen Chebotar, Xi~Chen, Krzysztof Choromanski, Tianli Ding, Danny Driess, Avinava Dubey, Chelsea Finn, et~al.
\newblock Rt-2: Vision-language-action models transfer web knowledge to robotic control.
\newblock \emph{arXiv preprint arXiv:2307.15818}, 2023.

\bibitem[Bu et~al.(2024)Bu, Zeng, Chen, Yang, Zhou, Yan, Luo, Cui, Ma, and Li]{bu2024closed}
Qingwen Bu, Jia Zeng, Li~Chen, Yanchao Yang, Guyue Zhou, Junchi Yan, Ping Luo, Heming Cui, Yi~Ma, and Hongyang Li.
\newblock Closed-loop visuomotor control with generative expectation for robotic manipulation.
\newblock \emph{arXiv preprint arXiv:2409.09016}, 2024.

\bibitem[Cen et~al.(2024)Cen, Wu, Liu, Yin, Pei, Yang, Chen, Duan, and Zhang]{cen2024using}
Jun Cen, Chenfei Wu, Xiao Liu, Shengming Yin, Yixuan Pei, Jinglong Yang, Qifeng Chen, Nan Duan, and Jianguo Zhang.
\newblock Using left and right brains together: Towards vision and language planning.
\newblock \emph{arXiv preprint arXiv:2402.10534}, 2024.

\bibitem[Cheang et~al.(2024)Cheang, Chen, Jing, Kong, Li, Li, Liu, Wu, Xu, Yang, et~al.]{cheang2024gr}
Chi-Lam Cheang, Guangzeng Chen, Ya~Jing, Tao Kong, Hang Li, Yifeng Li, Yuxiao Liu, Hongtao Wu, Jiafeng Xu, Yichu Yang, et~al.
\newblock Gr-2: A generative video-language-action model with web-scale knowledge for robot manipulation.
\newblock \emph{arXiv preprint arXiv:2410.06158}, 2024.

\bibitem[Chen et~al.(2025)Chen, Wu, Liu, Pan, Liu, Xie, Yu, and Ruan]{chen2025janus}
Xiaokang Chen, Zhiyu Wu, Xingchao Liu, Zizheng Pan, Wen Liu, Zhenda Xie, Xingkai Yu, and Chong Ruan.
\newblock Janus-pro: Unified multimodal understanding and generation with data and model scaling.
\newblock \emph{arXiv preprint arXiv:2501.17811}, 2025.

\bibitem[Chi et~al.(2023)Chi, Xu, Feng, Cousineau, Du, Burchfiel, Tedrake, and Song]{chi2023diffusion}
Cheng Chi, Zhenjia Xu, Siyuan Feng, Eric Cousineau, Yilun Du, Benjamin Burchfiel, Russ Tedrake, and Shuran Song.
\newblock Diffusion policy: Visuomotor policy learning via action diffusion.
\newblock \emph{The International Journal of Robotics Research}, page 02783649241273668, 2023.

\bibitem[Dosovitskiy et~al.(2020)Dosovitskiy, Beyer, Kolesnikov, Weissenborn, Zhai, Unterthiner, Dehghani, Minderer, Heigold, Gelly, et~al.]{dosovitskiy2020image}
Alexey Dosovitskiy, Lucas Beyer, Alexander Kolesnikov, Dirk Weissenborn, Xiaohua Zhai, Thomas Unterthiner, Mostafa Dehghani, Matthias Minderer, Georg Heigold, Sylvain Gelly, et~al.
\newblock An image is worth 16x16 words: Transformers for image recognition at scale.
\newblock \emph{arXiv preprint arXiv:2010.11929}, 2020.

\bibitem[Du et~al.(2023)Du, Yang, Dai, Dai, Nachum, Tenenbaum, Schuurmans, and Abbeel]{du2023learning}
Yilun Du, Sherry Yang, Bo~Dai, Hanjun Dai, Ofir Nachum, Josh Tenenbaum, Dale Schuurmans, and Pieter Abbeel.
\newblock Learning universal policies via text-guided video generation.
\newblock \emph{Advances in neural information processing systems}, 36:\penalty0 9156--9172, 2023.

\bibitem[Esser et~al.(2021)Esser, Rombach, and Ommer]{esser2021taming}
Patrick Esser, Robin Rombach, and Bjorn Ommer.
\newblock Taming transformers for high-resolution image synthesis.
\newblock In \emph{Proceedings of the IEEE/CVF conference on computer vision and pattern recognition}, pages 12873--12883, 2021.

\bibitem[Gafni et~al.(2022)Gafni, Polyak, Ashual, Sheynin, Parikh, and Taigman]{gafni2022make}
Oran Gafni, Adam Polyak, Oron Ashual, Shelly Sheynin, Devi Parikh, and Yaniv Taigman.
\newblock Make-a-scene: Scene-based text-to-image generation with human priors.
\newblock In \emph{European Conference on Computer Vision}, pages 89--106. Springer, 2022.

\bibitem[Ha and Schmidhuber(2018)]{ha2018world}
David Ha and J{\"u}rgen Schmidhuber.
\newblock World models.
\newblock \emph{arXiv preprint arXiv:1803.10122}, 2018.

\bibitem[He et~al.(2025)He, Zhang, Lin, Xu, and Pan]{he2025pre}
Haoran He, Yang Zhang, Liang Lin, Zhongwen Xu, and Ling Pan.
\newblock Pre-trained video generative models as world simulators.
\newblock \emph{arXiv preprint arXiv:2502.07825}, 2025.

\bibitem[He et~al.(2016)He, Zhang, Ren, and Sun]{he2016deep}
Kaiming He, Xiangyu Zhang, Shaoqing Ren, and Jian Sun.
\newblock Deep residual learning for image recognition.
\newblock In \emph{Proceedings of the IEEE conference on computer vision and pattern recognition}, pages 770--778, 2016.

\bibitem[Hou et~al.(2024)Hou, Zhang, Xiong, Pu, Zhao, Tong, Qiao, Dai, and Chen]{hou2024diffusion}
Zhi Hou, Tianyi Zhang, Yuwen Xiong, Hengjun Pu, Chengyang Zhao, Ronglei Tong, Yu~Qiao, Jifeng Dai, and Yuntao Chen.
\newblock Diffusion transformer policy.
\newblock \emph{arXiv preprint arXiv:2410.15959}, 2024.

\bibitem[Huang et~al.(2023)Huang, Yong, Ma, Linghu, Li, Wang, Li, Zhu, Jia, and Huang]{huang2023embodied}
Jiangyong Huang, Silong Yong, Xiaojian Ma, Xiongkun Linghu, Puhao Li, Yan Wang, Qing Li, Song-Chun Zhu, Baoxiong Jia, and Siyuan Huang.
\newblock An embodied generalist agent in 3d world.
\newblock \emph{arXiv preprint arXiv:2311.12871}, 2023.

\bibitem[Intelligence et~al.(2025)Intelligence, Black, Brown, Darpinian, Dhabalia, Driess, Esmail, Equi, Finn, Fusai, et~al.]{intelligence2025pi_}
Physical Intelligence, Kevin Black, Noah Brown, James Darpinian, Karan Dhabalia, Danny Driess, Adnan Esmail, Michael Equi, Chelsea Finn, Niccolo Fusai, et~al.
\newblock pi0.5: a vision-language-action model with open-world generalization.
\newblock \emph{arXiv preprint arXiv:2504.16054}, 2025.

\bibitem[Kim et~al.(2024)Kim, Pertsch, Karamcheti, Xiao, Balakrishna, Nair, Rafailov, Foster, Lam, Sanketi, et~al.]{kim2024openvla}
Moo~Jin Kim, Karl Pertsch, Siddharth Karamcheti, Ted Xiao, Ashwin Balakrishna, Suraj Nair, Rafael Rafailov, Ethan Foster, Grace Lam, Pannag Sanketi, et~al.
\newblock Openvla: An open-source vision-language-action model.
\newblock \emph{arXiv preprint arXiv:2406.09246}, 2024.

\bibitem[Kim et~al.(2025)Kim, Finn, and Liang]{kim2025fine}
Moo~Jin Kim, Chelsea Finn, and Percy Liang.
\newblock Fine-tuning vision-language-action models: Optimizing speed and success.
\newblock \emph{arXiv preprint arXiv:2502.19645}, 2025.

\bibitem[Li et~al.(2024)Li, Zhang, Guo, Zhang, Li, Zhang, Zhang, Zhang, Li, Liu, et~al.]{li2024llava}
Bo~Li, Yuanhan Zhang, Dong Guo, Renrui Zhang, Feng Li, Hao Zhang, Kaichen Zhang, Peiyuan Zhang, Yanwei Li, Ziwei Liu, et~al.
\newblock Llava-onevision: Easy visual task transfer.
\newblock \emph{arXiv preprint arXiv:2408.03326}, 2024.

\bibitem[Li et~al.(2025)Li, Gao, Sadigh, and Song]{li2025uva}
Shuang Li, Yihuai Gao, Dorsa Sadigh, and Shuran Song.
\newblock Unified video action model.
\newblock In \emph{arxiv}, 2025.

\bibitem[Li et~al.(2023)Li, Liu, Zhang, Yu, Xu, Wu, Cheang, Jing, Zhang, Liu, et~al.]{li2023vision}
Xinghang Li, Minghuan Liu, Hanbo Zhang, Cunjun Yu, Jie Xu, Hongtao Wu, Chilam Cheang, Ya~Jing, Weinan Zhang, Huaping Liu, et~al.
\newblock Vision-language foundation models as effective robot imitators.
\newblock \emph{arXiv preprint arXiv:2311.01378}, 2023.

\bibitem[Liu et~al.(2023{\natexlab{a}})Liu, Zhu, Gao, Feng, Liu, Zhu, and Stone]{liu2023libero}
Bo~Liu, Yifeng Zhu, Chongkai Gao, Yihao Feng, Qiang Liu, Yuke Zhu, and Peter Stone.
\newblock Libero: Benchmarking knowledge transfer for lifelong robot learning.
\newblock \emph{Advances in Neural Information Processing Systems}, 36:\penalty0 44776--44791, 2023{\natexlab{a}}.

\bibitem[Liu et~al.(2023{\natexlab{b}})Liu, Li, Wu, and Lee]{liu2023visual}
Haotian Liu, Chunyuan Li, Qingyang Wu, and Yong~Jae Lee.
\newblock Visual instruction tuning.
\newblock \emph{Advances in neural information processing systems}, 36:\penalty0 34892--34916, 2023{\natexlab{b}}.

\bibitem[Pertsch et~al.(2025)Pertsch, Stachowicz, Ichter, Driess, Nair, Vuong, Mees, Finn, and Levine]{pertsch2025fast}
Karl Pertsch, Kyle Stachowicz, Brian Ichter, Danny Driess, Suraj Nair, Quan Vuong, Oier Mees, Chelsea Finn, and Sergey Levine.
\newblock Fast: Efficient action tokenization for vision-language-action models.
\newblock \emph{arXiv preprint arXiv:2501.09747}, 2025.

\bibitem[Pomerleau(1988)]{pomerleau1988alvinn}
Dean~A Pomerleau.
\newblock Alvinn: An autonomous land vehicle in a neural network.
\newblock \emph{Advances in neural information processing systems}, 1, 1988.

\bibitem[Radford et~al.(2021)Radford, Kim, Hallacy, Ramesh, Goh, Agarwal, Sastry, Askell, Mishkin, Clark, et~al.]{radford2021learning}
Alec Radford, Jong~Wook Kim, Chris Hallacy, Aditya Ramesh, Gabriel Goh, Sandhini Agarwal, Girish Sastry, Amanda Askell, Pamela Mishkin, Jack Clark, et~al.
\newblock Learning transferable visual models from natural language supervision.
\newblock In \emph{International conference on machine learning}, pages 8748--8763. PmLR, 2021.

\bibitem[Rumelhart et~al.(1986)Rumelhart, Hinton, and Williams]{rumelhart1986learning}
David~E Rumelhart, Geoffrey~E Hinton, and Ronald~J Williams.
\newblock Learning representations by back-propagating errors.
\newblock \emph{nature}, 323\penalty0 (6088):\penalty0 533--536, 1986.

\bibitem[Sennrich et~al.(2015)Sennrich, Haddow, and Birch]{sennrich2015neural}
Rico Sennrich, Barry Haddow, and Alexandra Birch.
\newblock Neural machine translation of rare words with subword units.
\newblock \emph{arXiv preprint arXiv:1508.07909}, 2015.

\bibitem[Team(2024)]{team2024chameleon}
Chameleon Team.
\newblock Chameleon: Mixed-modal early-fusion foundation models.
\newblock \emph{arXiv preprint arXiv:2405.09818}, 2024.

\bibitem[Team et~al.(2024)Team, Ghosh, Walke, Pertsch, Black, Mees, Dasari, Hejna, Kreiman, Xu, et~al.]{team2024octo}
Octo~Model Team, Dibya Ghosh, Homer Walke, Karl Pertsch, Kevin Black, Oier Mees, Sudeep Dasari, Joey Hejna, Tobias Kreiman, Charles Xu, et~al.
\newblock Octo: An open-source generalist robot policy.
\newblock \emph{arXiv preprint arXiv:2405.12213}, 2024.

\bibitem[Tian et~al.(2024)Tian, Yang, Zeng, Wang, Lin, Dong, and Pang]{tian2024predictive}
Yang Tian, Sizhe Yang, Jia Zeng, Ping Wang, Dahua Lin, Hao Dong, and Jiangmiao Pang.
\newblock Predictive inverse dynamics models are scalable learners for robotic manipulation.
\newblock \emph{arXiv preprint arXiv:2412.15109}, 2024.

\bibitem[Tong et~al.(2024)Tong, Fan, Zhu, Xiong, Chen, Sinha, Rabbat, LeCun, Xie, and Liu]{tong2024metamorph}
Shengbang Tong, David Fan, Jiachen Zhu, Yunyang Xiong, Xinlei Chen, Koustuv Sinha, Michael Rabbat, Yann LeCun, Saining Xie, and Zhuang Liu.
\newblock Metamorph: Multimodal understanding and generation via instruction tuning.
\newblock \emph{arXiv preprint arXiv:2412.14164}, 2024.

\bibitem[Wang et~al.(2024)Wang, Zhang, Luo, Sun, Cui, Wang, Zhang, Wang, Li, Yu, et~al.]{wang2024emu3}
Xinlong Wang, Xiaosong Zhang, Zhengxiong Luo, Quan Sun, Yufeng Cui, Jinsheng Wang, Fan Zhang, Yueze Wang, Zhen Li, Qiying Yu, et~al.
\newblock Emu3: Next-token prediction is all you need.
\newblock \emph{arXiv preprint arXiv:2409.18869}, 2024.

\bibitem[Wen et~al.(2024)Wen, Zhu, Zhu, Tang, Li, Zhou, Li, Liu, Peng, Shen, et~al.]{wen2024diffusion}
Junjie Wen, Minjie Zhu, Yichen Zhu, Zhibin Tang, Jinming Li, Zhongyi Zhou, Chengmeng Li, Xiaoyu Liu, Yaxin Peng, Chaomin Shen, et~al.
\newblock Diffusion-vla: Scaling robot foundation models via unified diffusion and autoregression.
\newblock \emph{arXiv preprint arXiv:2412.03293}, 2024.

\bibitem[Wen et~al.(2025)Wen, Zhu, Li, Zhu, Tang, Wu, Xu, Liu, Cheng, Shen, et~al.]{wen2025tinyvla}
Junjie Wen, Yichen Zhu, Jinming Li, Minjie Zhu, Zhibin Tang, Kun Wu, Zhiyuan Xu, Ning Liu, Ran Cheng, Chaomin Shen, et~al.
\newblock Tinyvla: Towards fast, data-efficient vision-language-action models for robotic manipulation.
\newblock \emph{IEEE Robotics and Automation Letters}, 2025.

\bibitem[Wu et~al.(2023)Wu, Jing, Cheang, Chen, Xu, Li, Liu, Li, and Kong]{wu2023unleashing}
Hongtao Wu, Ya~Jing, Chilam Cheang, Guangzeng Chen, Jiafeng Xu, Xinghang Li, Minghuan Liu, Hang Li, and Tao Kong.
\newblock Unleashing large-scale video generative pre-training for visual robot manipulation.
\newblock \emph{arXiv preprint arXiv:2312.13139}, 2023.

\bibitem[Wu et~al.(2025)Wu, Yin, Feng, He, Li, Hao, and Long]{wu2025ivideogpt}
Jialong Wu, Shaofeng Yin, Ningya Feng, Xu~He, Dong Li, Jianye Hao, and Mingsheng Long.
\newblock ivideogpt: Interactive videogpts are scalable world models.
\newblock \emph{Advances in Neural Information Processing Systems}, 37:\penalty0 68082--68119, 2025.

\bibitem[Yu et~al.(2023)Yu, Cheng, Sohn, Lezama, Zhang, Chang, Hauptmann, Yang, Hao, Essa, et~al.]{yu2023magvit}
Lijun Yu, Yong Cheng, Kihyuk Sohn, Jos{\'e} Lezama, Han Zhang, Huiwen Chang, Alexander~G Hauptmann, Ming-Hsuan Yang, Yuan Hao, Irfan Essa, et~al.
\newblock Magvit: Masked generative video transformer.
\newblock In \emph{Proceedings of the IEEE/CVF Conference on Computer Vision and Pattern Recognition}, pages 10459--10469, 2023.

\bibitem[Zhai et~al.(2023)Zhai, Mustafa, Kolesnikov, and Beyer]{zhai2023sigmoid}
Xiaohua Zhai, Basil Mustafa, Alexander Kolesnikov, and Lucas Beyer.
\newblock Sigmoid loss for language image pre-training.
\newblock In \emph{Proceedings of the IEEE/CVF international conference on computer vision}, pages 11975--11986, 2023.

\bibitem[Zhang et~al.(2025)Zhang, Li, Cheng, Hu, Yuan, Chen, Leng, Jiang, Zhang, Li, et~al.]{zhang2025videollama}
Boqiang Zhang, Kehan Li, Zesen Cheng, Zhiqiang Hu, Yuqian Yuan, Guanzheng Chen, Sicong Leng, Yuming Jiang, Hang Zhang, Xin Li, et~al.
\newblock Videollama 3: Frontier multimodal foundation models for image and video understanding.
\newblock \emph{arXiv preprint arXiv:2501.13106}, 2025.

\bibitem[Zhao et~al.(2023)Zhao, Kumar, Levine, and Finn]{zhao2023learning}
Tony~Z Zhao, Vikash Kumar, Sergey Levine, and Chelsea Finn.
\newblock Learning fine-grained bimanual manipulation with low-cost hardware.
\newblock \emph{arXiv preprint arXiv:2304.13705}, 2023.

\bibitem[Zhen et~al.(2024)Zhen, Qiu, Chen, Yang, Yan, Du, Hong, and Gan]{zhen20243d}
Haoyu Zhen, Xiaowen Qiu, Peihao Chen, Jincheng Yang, Xin Yan, Yilun Du, Yining Hong, and Chuang Gan.
\newblock 3d-vla: A 3d vision-language-action generative world model.
\newblock \emph{arXiv preprint arXiv:2403.09631}, 2024.

\bibitem[Zhou et~al.(2024)Zhou, Yu, Babu, Tirumala, Yasunaga, Shamis, Kahn, Ma, Zettlemoyer, and Levy]{zhou2024transfusion}
Chunting Zhou, Lili Yu, Arun Babu, Kushal Tirumala, Michihiro Yasunaga, Leonid Shamis, Jacob Kahn, Xuezhe Ma, Luke Zettlemoyer, and Omer Levy.
\newblock Transfusion: Predict the next token and diffuse images with one multi-modal model.
\newblock \emph{arXiv preprint arXiv:2408.11039}, 2024.

\end{thebibliography}


\end{document}